\documentclass[runningheads]{llncs}
\usepackage[T1]{fontenc}
\usepackage{graphicx}
\RequirePackage{hyperref}
\begin{document}
\title{On the detection of anomalous or out-of-distribution data in vision models using statistical techniques}
\titlerunning{On the detection of anomalous or out-of-distribution data}
%
\author{Laura O'Mahony\inst{1} \and
David JP O'Sullivan\inst{1} \and
Nikola S. Nikolov\inst{1}}
\authorrunning{L. O'Mahony et al.}
%
\institute{University of Limerick, Castletroy, Limerick, V94 T9PX, Ireland \\
\email{\{lauraa.omahony,david.osullivan,nikola.nikolov\}@ul.ie}}
\maketitle              
\begin{abstract}
Out-of-distribution data and anomalous inputs are vulnerabilities of machine learning systems today, often causing systems to make incorrect predictions. 
  The diverse range of data on which these models are used makes detecting atypical inputs a difficult and important task. 
  We assess a tool, Benford's law, as a method used 
  to quantify the difference between real and corrupted inputs. 
  We believe that in many settings, it could function as a filter for anomalous data points and for signalling out-of-distribution data. 
  We hope to open a discussion on these applications and further areas where this technique is underexplored.

\end{abstract}

\section{Introduction}\label{sec1}

Deploying machine learning systems in the real world brings with it multiple AI safety challenges, for example, issues related to model robustness and monitoring.  
Research on the detection of adversarial examples, anomalous inputs, and flagging when the data a model is deployed on is out-of-distribution attempts to tackle such issues. 
Out-of-distribution means there is a difference in distributional properties between the test, training, and real world data. 
For machine learning models, even simple shifts in data distribution can lead to a large drop in performance~\cite{koh2021wilds}. 
An example of a failure caused by data anomalies is the discovery that classifiers often fail by giving high-confidence predictions on anomalous inputs when their predictions are incorrect~\cite{nguyen2015deep}. The failure of such classifiers to indicate where they are likely wrong is problematic~\cite{amodei2016concrete}. 
In this work, we study how a range of image corruptions appreciably deviate in distribution from natural images by leveraging a dataset containing various image corruption types.  
Our work trials the broad applicability of a method that could act as a filter for anomalous inputs and indicate distribution shifts. We find the method  reflects some corruption types in the experiments performed and thus may be practical as an accompaniment to other methods for out-of-distribution data and anomalous input detection.

A vast amount of literature exists aiming to detect out-of-distribution and anomalous data \cite{yang2021generalized}. 
Some of these methods focus on analysing the activations of hidden layers \cite{stano2020explaining,henzinger2019outside,hashemi2021gaussian,olber2022detection}. 
Many works formulate the problem as a model selection between an in-distribution and an out-of-distribution density model and use a likelihood-ratio based approach to detect out-of-distribution data~\cite{bishop1994novelty,schirrmeister2020understanding,zhang2022out}. 
Another approach relates to the softmax distribution~\cite{hendrycks2016baseline}. 
Other authors utilise early-layer outputs~\cite{abdelzad2019detecting} or gradient space in out-of-distribution data detection \cite{huang2021importance}. A technique known as outlier exposure tackles classifier overconfidence by training anomaly detectors against an auxiliary outlier dataset~\cite{hendrycks2018deep}. 
There also exists previous literature concerning the properties, such as the mean, of a dataset computed in the presence of outliers~\cite{steinhardt2017resilience}. 
Other work on the low-order statistics of data approaches the problem of out-of-distribution data by analysing the low-order statistics and framing it as a statistical hypothesis testing problem~\cite{haroush2021statistical}. 

Our approach differs in that it uses a less-known property of image distributions relating to Benford's law (BFL), which has a proven track record in image forensics, as is discussed in Sec.~\ref{background}. 
To our knowledge, the use of Benford's law is novel in the context of out-of-distribution and anomaly detection in the context of machine learning classification. 
We note that our approach is compatible with other differing out-of-distribution heuristics and anomaly detection techniques, meaning it could complement other methods in the literature. 
Furthermore, our approach is surprisingly simple and computationally inexpensive as it does not modify the model parameters or architecture, making it scalable.

The paper is structured as follows: 
In Sec.~\ref{background}, we outline the theory and current applications for the distributional summaries of images. In Sec.~\ref{analysis}, we outline the pipeline used to measure an image's deviation from its theoretical distribution before introducing the dataset used to evaluate this pipeline. We then present the experimental results. We also discuss possible extensions to this work and its applications before concluding in Sec.~\ref{conclusion}.

\section{Background} \label{background}

In this section, we introduce the Benford equation as a benchmark that can be used to compare the statistical distributional properties of images. 
In 1938, Benford~\cite{benford1938law} introduced BFL for numbers. 
This paper predicts the statistical frequency of the leading digits (LDs) for a set of measurements of a broad range of ``natural'' quantities~\cite{raimi1976first}. 
Here, natural data is understood to be data collected by observing natural phenomena. 
It is worth noting that the cases where the rule applies are often determined experimentally rather than mathematically~\cite{crișan2021analyzing}. 
Benford's law has been empirically observed for many natural quantities, such as the lengths of rivers, population numbers, stock market prices, etc. The predicted distribution, $p(d)$ is given by Eq. \ref{eq:Benford_eq}, where $d$ is the LD of the number (e.g., `245326' has the LD `2'. The remaining digits `45326' are discarded as Benford's law only concerns the distribution of the LDs.),

\setlength{\belowdisplayshortskip}{\belowdisplayskip}
\begin{equation}
    p(d) = log_b\left(1+\frac{1}{d}\right), \hspace{3mm} d = 1,2,...,b-1.\label{eq:Benford_eq}
\end{equation}

\begin{figure}[!htbp]
\centering
\includegraphics[width=0.8\textwidth]{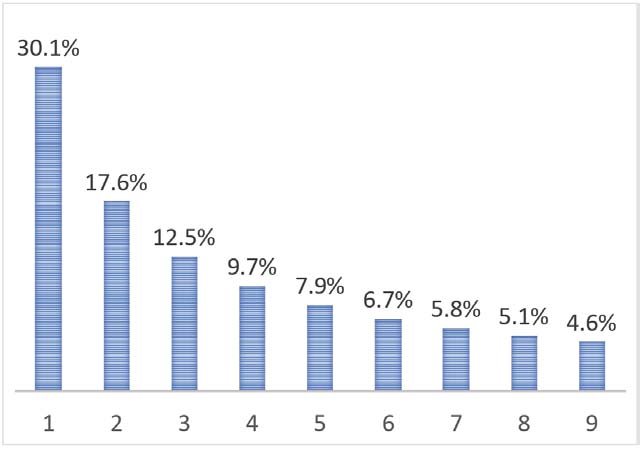}
\caption{\label{fig:Benfords_law}The theoretical distribution of first digits in data that obeys Benford's law. The height of the bar is the percentage of numbers that start with the digit.}
\end{figure}

As described in the equation, the nature of probabilities of the first digits is logarithmic to the base $b$, for invariant $b$. 
The distribution given by the Benford equation, i.e., Eq. \ref{eq:Benford_eq}, in base 10, is shown in Fig. \ref{fig:Benfords_law}. 
As shown in this figure, the occurrence of the smaller first digits is more frequent than larger first digits.

It has been observed that this rule is not well-fitted by the LD statistics of altered data~\cite{diekmann2007not}.
As a result, it has been useful for purposes such as fraud detection in taxing and accounting~\cite{durtschi2004effective,asllani2015using}. 
It has also been used to detect anomalies that require investigation for fraud in customs declarations~\cite{cerioli2019newcomb}.

BFL can also be applied to image data, provided an appropriate transform is made before the distribution is extracted from images. 
The discrete cosine transform (DCT)~\cite{ahmed1974discrete} is a widely used transformation technique in signal processing and data compression~\cite{rao2014discrete}. 
It expresses a finite sequence of data points as a sum of cosine functions oscillating at various frequencies. 
It has been shown that the DCT coefficients of ``natural'' images can typically be modelled by a Laplacian-like distribution, which follows BFL~\cite{lam2000mathematical}. 
Therefore, images mapped into the DCT domain typically follow BFL closely~\cite{perez2007benford}, giving us a transform that can be used to extract a distribution that should follow BFL. 
Previous work has exploited this for applications such as detecting JPEG compression~\cite{pasquini2014multiple,milani2014discriminating}, synthetic imagery~\cite{acebo2005benford}, estimating the amount of processing that has been applied to an image, and detecting image tampering~\cite{singh2015analysis}. 
It has been applied to detect if an image is generated by a generative adversarial network (GAN)~\cite{bonettini2021use}. 
It has also been shown that images that reflect only a small part of a scene do not fit BFL well~\cite{acebo2005benford}. 
A connection between a neural network's generalisation capability and a BFL related metric has been studied~\cite{sahu2021rethinking}.

The extensive analysis of BFL in image forensics and otherwise lends the question of whether the rule could be employed in AI safety applications,  
such as detecting anomalies and indicating if data is out-of-distribution. 
We ask if this theoretical structure of natural images shifts as images become less legible, allowing us to detect anomalies and reflect when data is out-of-distribution. 


\section{Analysis of image discrete cosine transform coefficients} \label{analysis}

\subsection{Methodology for extracting a distributional comparison measure from images} \label{set-up}

In this section, we provide the details of how images are processed to analyse the DCT coefficient distribution of images and how we measure the divergence of the empirical distribution of images from the BFL theoretical distribution.

\begin{figure}[!htb]
\centering
\includegraphics[width=1\textwidth]{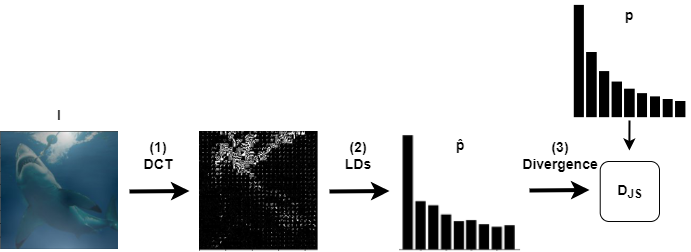}
\caption{The procedure of quantifying the difference between the distribution of the image discrete cosine transform (DCT) coefficients and the theoretical distribution given by Benford's law consists of (1) applying a discrete cosine transform to the image $I$, followed by (2) calculating the leading digit (LD) distribution of the transform coefficients,  
and finally (3) a statistical comparison between the theoretical distribution $p$ and observed distribution $\hat{p}$ of the transformed coefficients is made.
The raw image $I$ is taken from the ImageNet-C dataset~\cite{hendrycks2018benchmarking}.}\label{fig:pipeline}
\end{figure}

We first need to obtain a distributional representation of an image so that we can compare it to the theoretical distribution. 
The process of obtaining this distribution and measuring the divergence from BFL is depicted in Fig.~\ref{fig:pipeline}. 
Given an RGB image $I$, the image is initially converted to greyscale. 
The image is then divided into 8x8 pixel blocks that do not overlap. Each block is transformed using the DCT. 
The empirical distribution of the LDs of the DCT coefficients from each block is calculated with respect to a base, e.g., base 10. 
This outputs a probability mass function (pmf), $\hat{p}(d)$, for the image in question, $I$. This computed pmf is compared to the theoretical distribution, given by the Benford equation, i.e., Eq. \ref{eq:Benford_eq}. 
A statistical metric, the Jensen-Shannon divergence~\cite{lin1991divergence}, is employed to investigate the difference between these distributions. 
This gives us $D_{JS}$, a measure of the distributional difference between the empirical and theoretical distribution of $I$.

This process provides a measure of how far the distribution of an image is from BFL. We can then use this as a comparison between clean images and images with different corruption types applied at varying severity levels, which we could think of as distributional shifts in the data. To test this process, we use the Image-C dataset~\cite{hendrycks2018benchmarking}, which we describe in the next section.

\subsection{Data}

To test our pipeline described in Sec.~\ref{set-up}, the ImageNet-C dataset was repurposed to assess whether varying levels of image corruption are reflected by deviations of images from BFL.

\begin{figure}[!htb]
\centering
\includegraphics[width=0.9\textwidth]{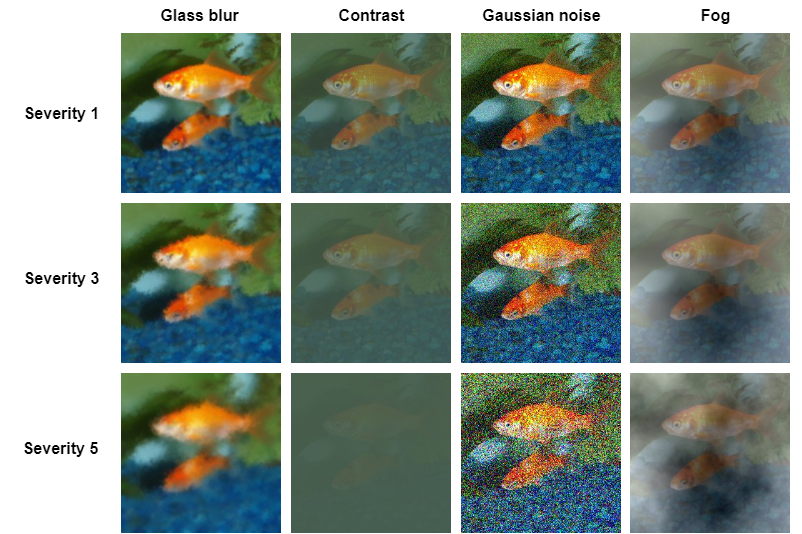}
\caption{\label{fig:ImageNet-C}An example of images taken from ImageNet-C~\cite{hendrycks2018benchmarking} representing different corruption types.}
\end{figure}

ImageNet-C
\footnote{License \href{https://creativecommons.org/licenses/by/4.0/legalcode}{Creative Commons Attribution 4.0 Intern}, DOI:
10.5281/zenodo.2235448ational}
\cite{hendrycks2018benchmarking} is a dataset created by Hendrycks and Dietterich, which is designed to establish benchmarks for image classifier robustness. 
The dataset contains 15 distinct types of image corruptions (noise, blur, weather, and digital corruptions). 
Each of these corruption types has severity levels 1 - 5, where higher levels of severity indicate a higher level of corruption being applied. 
The dataset is algorithmically generated from the ImageNet validation data~\cite{ILSVRC15} and intended to evaluate the performance of a model on common image corruptions.
Therefore, the data is the standard ImageNet size and consists of 1000 ImageNet classes. 
Both the type and severity of corruption give us a unique dataset to test if we can detect differences in the distribution of images using our method. 
See Fig.~\ref{fig:ImageNet-C} for an example of an image with a subset of the corruption types applied at corruption severity levels 1, 3 and 5 respectively.
Four different corruption types are shown, corresponding to a common image from the ImageNet validation data.

\subsection{Results of the image comparison metric and discussion}

\begin{figure}[!htbp] 
\centering
\includegraphics[width=0.9\textwidth]{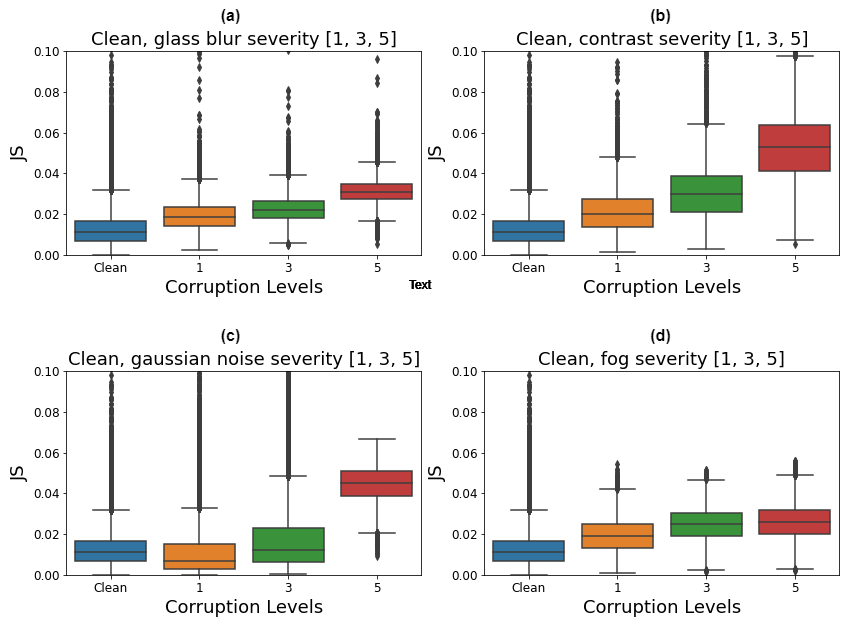}
\caption{\label{fig:subcategory_JS_DCT}The Jensen-Shannon divergence of the statistics of the discrete cosine transform coefficients for clean ImageNet images (blue), images of corruption severity 1 (orange), corruption severity 3 (green), and corruption severity 5 (red).}
\end{figure}

The pipeline detailed in Sec.~\ref{set-up} was applied to the 50,000 ImageNet validation dataset~\cite{ILSVRC15} images.
Due to computational constraints, it was applied to the ImageNet-C images of corruption severity levels 1, 3 and 5, corruption types glass blur, contrast, Gaussian noise, and fog
\footnote{Code available at \url{https:/github.com/lomahony/first-digits}.}. 
We first look at the distribution of these divergence statistics. 
From Fig.~\ref{fig:subcategory_JS_DCT}, we can see that, typically, across all corruption types, with the exception of Gaussian noise, corrupted images show an increase in the median divergence from the Benford's law (BFL) benchmark compared to uncorrupted images.
Higher severity levels of corruption types glass blur and contrast lead to a larger median divergence from the BFL benchmark compared to lower severity levels of corruption. 
A different pattern is seen for the Gaussian noise corrupted images at severity levels 1 and 3. 
A possible explanation is that Gaussian noise may interfere with the Laplacian-like distribution followed by the DCT coefficients of images. 

\begin{figure}[!htbp]
\centering
\includegraphics[width=0.8\textwidth]{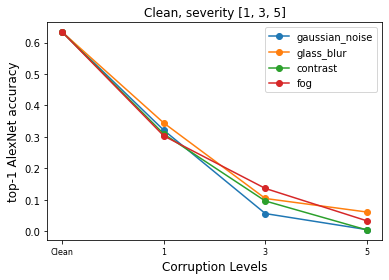}
\caption{\label{fig:accuracies}The top-1 predictive accuracy evaluated using AlexNet \cite{krizhevsky2017imagenet} is shown for clean ImageNet images, images of corruption severity 1, corruption severity 3, and corruption severity 5.}
\end{figure}

Further experiments assessed whether the proposed approach can capture model degradation on out-of-distribution data. A popular pretrained image classification model called AlexNet \cite{krizhevsky2017imagenet} was used as a benchmark model to compare the test accuracy of the top-1 prediction accuracy at the selected corruption levels. 
Fig.~\ref{fig:accuracies} shows that, as expected, the baseline ImageNet validation accuracy of AlexNet of 0.633 decreased as the level of corruption increased for each corruption category. 
In some cases, the model performance drop did not reflect a higher median divergence from the BFL benchmark, such as the median Jensen-Shannon divergence for Gaussian noise corrupted images at severity levels 1 and 3 compared with the median divergence for the clean ImageNet images. 
In other cases, a considerable jump in the median divergence from the BFL benchmark reflected a large model performance degradation. 
An example of this is the jump in Jensen-Shannon divergence between contrast and Gaussian noise corrupted images at severity levels 3 and 5. 
 This was reflected by a very low top-1 accuracy. 
From these results, a clear limitation of this technique is that the distributional shift may impact model performance significantly more than what is reflected by the divergence statistic. 
This suggests this approach may be useful as a complementary approach but should not replace other anomaly or out-of-distribution detection methods. 
Another limitation of this approach is that it may be useful for sensory anomaly detection (detecting changes in the input space e.g., indicating where samples may be from a different domain), but not semantic anomaly detection (detecting changes in the label space e.g., detecting the occurrence of new classes) \cite{yang2021generalized}. 

A generalisation of BFL has been shown to better fit natural images due to the properties of the DCT transform~\cite{perez2007benford}. 
This suggests an extension of this work could involve investigating if a generalised Benford's equation improves on the results presented here for Gaussian noise as well as the other corruption types. 
In addition, further extensions of this work could involve calculating the empirical distribution of training images. This could be used as the benchmark instead of Benford's equation to test if this technique could be used outside of natural images. 
As BFL is applicable to various data types, another potential future direction could involve testing the methodology on other data types. 
Further experiments could also involve testing to what extent model performance drop correlates with the level of distribution shift detected using our pipeline.

\section{Conclusion} \label{conclusion}

In this paper, we investigated if the divergence of image statistics from Benford's Law is reflected when images are corrupted at various levels. 
Importantly, we do not claim that this phenomenon is universal, but we found there to be a difference in image distribution across many corruption types for natural images taken from ImageNet. 
Results show that for many corruption types, images that are corrupted to a higher level typically deviate from the expected distribution more. 
This positive result motivates further testing, and we believe extensions of the methods employed in this paper are likely to give positive results. 
Future work might involve testing the extent that the proposed approach reflects the model degradation on out-of-distribution data. 
This technique could be added to the toolkit as a low computational filter for anomalous or out-of-distribution data or to assess the reliability of images before attempting to classify them. 
It could even be investigated as a coarse filter to detect some types of adversarial examples. 
Furthermore, wider ramifications of the successful differentiation in distribution between clean and corrupted images that the authors are not aware of may lead to further interesting research questions.

\subsubsection{Acknowledgements} This publication has emanated from research conducted with the financial support of Science Foundation Ireland under Grant Number 18/CRT/6049. For the purpose of Open Access, the authors have applied a CC BY public copyright licence to any Author Accepted Manuscript version arising from this submission. 

%
%
%
\bibliographystyle{splncs04}
\bibliography{bibliography}
\end{document}